%% file: acl_latex.tex
\pdfoutput=1

\documentclass[11pt]{article}

\usepackage{latex/acl}

\usepackage{times}
\usepackage{latexsym}

\usepackage[T1]{fontenc}

\usepackage[utf8]{inputenc}

\usepackage{microtype}

\usepackage{inconsolata}

\usepackage{graphicx}
\usepackage{threeparttable}
\usepackage{booktabs}
\usepackage{multirow}
\usepackage{pifont}
\usepackage{enumitem}
\usepackage[most]{tcolorbox}
\usepackage{listings}
\newcommand{\e}[1]{{$#1$}}

\usepackage{authblk}

\input{macro}

\title{Augment before You Try: Knowledge-Enhanced Table Question \\Answering via Table Expansion}

\author[1]{\textbf{Yujian Liu}}
\author[1]{\textbf{Jiabao Ji}}
\author[2]{\textbf{Tong Yu}}
\author[2]{\textbf{Ryan Rossi}}
\author[2]{\\\textbf{Sungchul Kim}}
\author[2]{\textbf{Handong Zhao}}
\author[2]{\textbf{Ritwik Sinha}}
\author[3]{\textbf{Yang Zhang}}
\author[1]{\textbf{Shiyu Chang}}

\affil[ ]{$^1$UC Santa Barbara, $^2$Adobe Research, $^3$MIT-IBM Watson AI Lab}

\affil[ ]{$^1${\small\{\texttt{yujianliu,jiabaoji,chang87\}@ucsb.edu},} $^3${\small \texttt{yangzhan@mit.edu}}}
\affil[2]{{\small\{\texttt{tyu,ryrossi,sukim,hazhao,risinha\}@adobe.com}}}

\begin{document}
\maketitle
\begin{abstract}
\input{sections/00_abstract}
\end{abstract}

\section{Introduction}
\input{sections/01_intro}

\section{Related work}
\input{sections/02_relatedwork}

\section{Methodology}
\input{sections/03_method}

\section{Experiments}
\input{sections/04_experiment}

\section{Conclusion}
\input{sections/05_conclusion}

\section{Limitation}
\input{sections/limitations}

\bibliography{custom}

\clearpage
\appendix

\input{sections/06_appendix}

\end{document}

%% file: macro.tex
\makeatletter
\lst@InstallKeywords k{attributes}{attributestyle}\slshape{attributestyle}{}ld
\makeatother

\lstdefinestyle{demo}{
    basicstyle=\fontsize{8}{9}\ttfamily,
    keywordstyle=\color{blue},
    commentstyle=\color{gray},
    stringstyle=\color{green},
    showstringspaces=false,
    breaklines=true,
    breakatwhitespace=false,
    breakindent=0pt,
    escapeinside={(*@}{@*)},
    literate={á}{{\'a}}1 {ã}{{\~a}}1 {é}{{\'e}}1,
}

\lstdefinestyle{text}{
    basicstyle=\fontsize{8}{10}\ttfamily,
    showstringspaces=false,
    breaklines=true,
    breakatwhitespace=false,
    breakindent=0pt,
    keepspaces=true,
    showspaces=false,   
    escapeinside={(*@}{@*)}
}

%% file: sections/00_abstract.tex
Table question answering is a popular task that assesses a model's ability to understand and interact with structured data. However, the given table often does not contain sufficient information for answering the question, necessitating the integration of external knowledge. Existing methods either convert both the table and external knowledge into text, which neglects the structured nature of the table; or they embed queries for external sources in the interaction with the table, which complicates the process. In this paper, we propose a simple yet effective method to integrate external information in a given table. Our method first constructs an augmenting table containing the missing information and then generates a SQL query over the two tables to answer the question. Experiments show that our method outperforms strong baselines on three table QA benchmarks. Our code is publicly available at \url{https://github.com/UCSB-NLP-Chang/Augment_tableQA}.

%% file: sections/01_intro.tex
\begin{figure*}[t]
    \centering
    \includegraphics[width=0.9\linewidth]{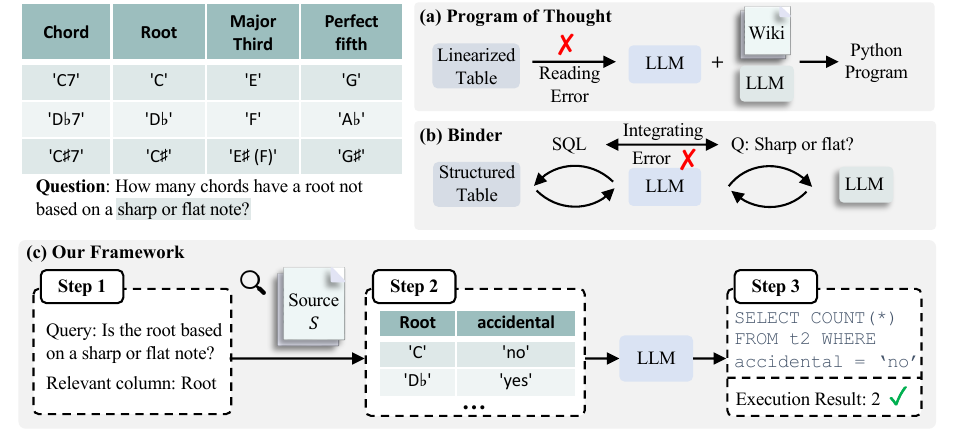}
    \caption{Comparison between \texttt{Program-of-Thought}, \texttt{Binder}, and our method.}
    \label{fig:intro-example}
    \vspace{-5mm}

\end{figure*}

Tables are ubiquitous types of information sources that have attracted significant attention in the NLP community. Researchers have developed models that understand tabular data and perform various tasks, including table question answering (QA) \citep{pasupat-liang-2015-compositional, chen-etal-2020-hybridqa, nan-etal-2022-fetaqa, zhu-etal-2021-tat, chen-etal-2021-finqa}, table fact verification \citep{Chen2020TabFact:, aly-etal-2021-fact}, table-to-text generation \citep{parikh-etal-2020-totto, chen-etal-2020-logical, nan-etal-2021-dart}, \emph{etc.} A critical challenge in these tasks is that tables often lack sufficient information for the task at hand, which necessitates the integration of additional knowledge. For example, in Figure \ref{fig:intro-example}, to answer the question \textit{`How many chords have a root not based on a sharp or flat note?'}, a model needs to have the knowledge of whether each root is based on a sharp or flat note, which is not provided in the table and can only be obtained from external sources.

Existing methods for integrating information from tables and external sources can be mainly categorized into two groups. The first method, exemplified by \texttt{Program-of-Thought} \citep{chen2023program}, linearizes the table into text and combines it with external knowledge in textual format \citep{xie-etal-2022-unifiedskg, chen-2023-large}. However, the linearized table no longer has the structured format, making it difficult to retrieve required values from the table and perform comparisons and calculations.

An alternative approach, represented by \texttt{Binder} \citep{cheng2023binding}, combines the symbolic language execution with large language models (LLMs). In this method, the model interacts with the table through symbolic language like SQL, which maintains the structured format. Part of the SQL query is replaced with an LLM query that extracts knowledge from the LLM and returns the results for further SQL execution. For instance, in Figure \ref{fig:intro-example} (b), the method queries LLMs for whether each root is sharp or flat and uses the results as a filtering criterion in a SQL statement. However, it requires the model to learn to embed LLM queries in the standard SQL language, which differs substantially from the SQL statements the model has been trained on. As a result, it is more likely to generate syntactically wrong statements that lead to execution errors.

In this paper, we propose a simple yet effective method for combining external knowledge with a given table. As shown in Figure \ref{fig:intro-example} (c), our method starts by analyzing the additional information required for answering the question. It then queries a knowledge source for the information and organizes the results in a tabular format. This newly created table \textit{augments} the original table with additional information, and a SQL query is generated to obtain the answer from the two tables. Such an augment-then-generate pipeline eliminates the need to embed LLM queries in SQL statements while preserving the structured format of the table.

We evaluate our method on three table QA datasets that require different types of external knowledge \citep{chen-etal-2021-finqa, zhu2021tat, pasupat-liang-2015-compositional}. Our method outperforms or matches strong baselines on all datasets. Particularly, it demonstrates significant improvements over \texttt{Program-of-Thought} in questions with large tables or require complex tabular operations, and compared to \texttt{Binder}, it exhibits fewer execution errors and achieves better performance (8.32$\%$).

%% file: sections/02_relatedwork.tex
\label{sec:related-work}

\textbf{Table question answering} task requires both the ability to reason over structured data and to understand textual contents in the table. Traditional methods utilize semantic parsing to convert the natural language question into executable commands, which retrieve and process data in the table to obtain answers \citep{zettlemoyer2012learning, berant2013semantic, yin2017syntactic, zhong2017seq2sql, shaw2020compositional, yu-etal-2018-spider, rajkumar2022evaluating}. However, these methods require all question-related information to present in the table in a rigorous format, which is limited when applied to web tables that often do not have a clean schema. Recent works pre-train neural models on large-scale tabular data, and directly encode tables and generate answers in an end-to-end fashion \citep{liu2022tapex, xie-etal-2022-unifiedskg, herzig-etal-2020-tapas, yin-etal-2020-tabert, zhao2022reastap, deng2020turl}. To reduce the training cost, some works further leverage LLMs to read and reason over tables \citep{chen-2023-large, pourreza2023dinsql}.

Although end-to-end methods achieve impressive performance on table QA benchmarks, their predictions lack interpretability and are not robust to input perturbations \citep{yang-etal-2022-tableformer}. For this reason, recent works propose to combine LLMs with symbolic language execution. Particularly, \citet{cheng2023binding} expands the SQL language by incorporating function calls to LLMs in SQL statements. \citet{ye2023large} utilizes LLMs to decompose the question and table into sub-problems that can be solved with SQL queries. \citet{chen2023program} prompts LLMs to generate the reasoning process as Python programs. Similar to our method, a concurrent work \citep{wang2024chainoftable} proposes to dynamically update the table content in the reasoning process. They employ LLMs to iteratively generate operations such as selecting a subset of rows or adding a new column, and the final resulting table is fed to LLMs to generate the answer. However, their chain of operations is prone to error propagation, while our method retains the original table content and augments it with required information.

%% file: sections/03_method.tex
\subsection{Problem Formulation}
Given a natural language question \e{Q}, a table \e{T}, and a knowledge source \e{S}, the task is to generate a correct answer for the question. Crucially, \e{T} might not contain all the necessary information to answer the question, which necessitates the use of \e{S} to obtain additional information. In this paper, we consider \e{S} to be either a relevant text document or an LLM that we can query.

\subsection{Overall Framework}
Our method contains three steps, as illustrated in Figure \ref{fig:intro-example} (c). The detailed instructions and examples for each step are listed in Appendix \ref{app:imp-details}.

\vspace{0.05in}
\noindent
\textbf{Step 1: Analyze question.}\quad
An LLM is instructed to analyze the given question and table to determine what additional information is needed to answer the question. We instruct the LLM to first list out all the necessary information for answering \e{Q}. For each piece of information, it then determines if the information is present in \e{T} or not. The output of this step is a list of queries that can be later used to obtain additional information from \e{S}, or empty if no additional information is needed. For example, in Figure \ref{fig:intro-example} (c), the model outputs \textit{`Is the root based on a sharp or flat note?'}.
Additionally, for information that needs to be obtained based on the table, the LLM will also specify which columns are needed, \emph{e.g.,} the model specifies that the query needs to be answered for each row in the `Root' column.

\vspace{0.05in}
\noindent
\textbf{Step 2: Construct augmenting table.}\quad
Using output queries from step 1, the LLM is used to obtain corresponding information from the source \e{S}. Specifically, when \e{S} is a text document, this step is similar to the reading comprehension task where the LLM needs to extract answers to the queries from the document. When \e{S} is an LLM, this step resembles a QA task where the LLM needs to directly answer the query. Finally, the obtained information is organized into a separate table that can complement the existing table \e{T}. Figure \ref{fig:intro-example} (c) shows an example where a new table of two columns is constructed. It is worth mentioning that this step is flexible and can be easily extended to other types of sources \e{S}.

\vspace{0.05in}
\noindent
\textbf{Step 3: Generate SQL query.}\quad
With the original and newly constructed tables, the LLM then generates a SQL query that can be executed to obtain the answer to the question. Importantly, the two tables contain sufficient information for answering \e{Q}, and the LLM can generate a standard SQL query, which is easier and more similar to its pre-training data.

%% file: sections/04_experiment.tex
In this section, we empirically evaluate our method on table QA benchmarks, focusing on two types of questions that might require external knowledge from different sources.
\begin{itemize}[leftmargin=*,noitemsep,topsep=0.2pt]
    \item \textbf{Open-domain knowledge} where external information comes from an open domain. We use the embedded knowledge in LLMs as the source.
    \item \textbf{Closed-domain knowledge} where all information is within a given table and a text document. In this case, the document is the external source.
\end{itemize}
We will discuss the common experiment settings in Section \ref{subsec:setup} and individual experiments for each type in Sections \ref{subsec:exp-open} and \ref{subsec:exp-closed} respectively.

\subsection{Experiments Setup}
\label{subsec:setup}

\paragraph{Implementation details.}
We prompt an LLM with detailed instructions and several in-context examples to complete all three steps in our method. To feed the table to the LLM for question analysis (Step 1) and generating SQL queries (Step 3), we linearize the table by concatenating columns with special tokens (\emph{e.g.,} `|') following previous works \citep{chen2023program}. In all experiments, we use \texttt{GPT-3.5-turbo-1106} through the official API as the backbone LLM for our method and all baselines. We use greedy decoding (\emph{i.e.,} temperature equals 0) for our method and all baselines. To have a fair comparison, we use the same number of in-context examples as baselines (details in Appendix \ref{app:imp-details}).

\paragraph{Baselines.}
We compare with five LLM-based baselines. \ding{182} \texttt{End-to-End} that prompts the LLM to directly output the answer given the table, question, and optionally the text document. \ding{183} \texttt{Table-CoT} \citep{chen-2023-large} that uses the chain-of-thought prompting \citep{wei2022chain} to ask the model to additionally output the reasoning chain. \ding{184} \texttt{Dater} \citep{ye2023large}, \ding{185} \texttt{Binder} \citep{cheng2023binding}, and \ding{186} \texttt{Program-of-Thought} (\texttt{PoT}) \citep{chen2023program} that combine LLMs with symbolic language execution. Please refer to Section \ref{sec:related-work} for details. Particularly, since \texttt{Binder} does not generate the reasoning chain, we include an improved variant where we add chain-of-thought prompting (\texttt{Binder+CoT}).

\paragraph{Evaluations.}
We adopt the exact match rate (EM) between the model-predicted answer and the ground-truth answer as the metric. We use the same evaluation code for all methods for fair comparison.

\subsection{Open-Domain Knowledge}
\label{subsec:exp-open}

\paragraph{Datasets.}
For open-domain knowledge, we evaluate on \textsc{WikiTQ} dataset \citep{pasupat-liang-2015-compositional}, which requires complex table reasoning to answer the question. According to \citet{shi-etal-2020-potential}, around 20\% of questions in \textsc{WikiTQ} are not answerable by pure SQL queries, which are likely to require additional knowledge not present in the table. We evaluate all methods on a subset of 1000 samples in the test set due to the cost of full evaluation.

\begin{figure}
    \centering
    \includegraphics[width=0.85\linewidth]{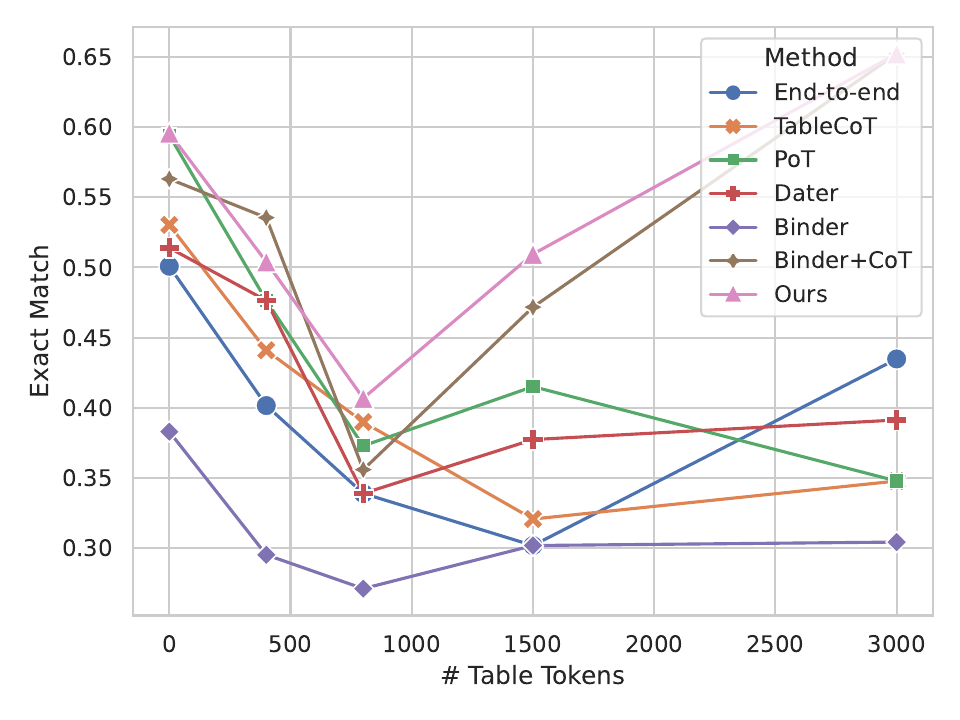}
    \vspace{-3mm}
    \caption{Performance grouped by table length.}
    \label{fig:wikitq-length}
    \vspace{-3mm}
\end{figure}

\paragraph{Results.}
Table \ref{tab:wikitq} presents the EM. There are two observations from the table. First, methods that involve program execution are generally better than those that do not, highlighting the value of accurate data retrieval or processing. Second, our method achieves the best performance, showing its effectiveness. To further evaluate scalability across table sizes, Figure \ref{fig:wikitq-length} plots the performance breakdown by the number of tokens in the table. As can be observed, our method and \texttt{Binder+CoT} are the only methods that maintain performance on large tables, whereas methods that rely on LLMs to extract information from linearized tables such as \texttt{Table-CoT} and \texttt{PoT} suffer significant performance degradation on large tables. This illustrates the advantage of SQL queries when interacting with the table.

\input{tables/wikitq}

\begin{figure}
    \centering
    \includegraphics[width=\linewidth]{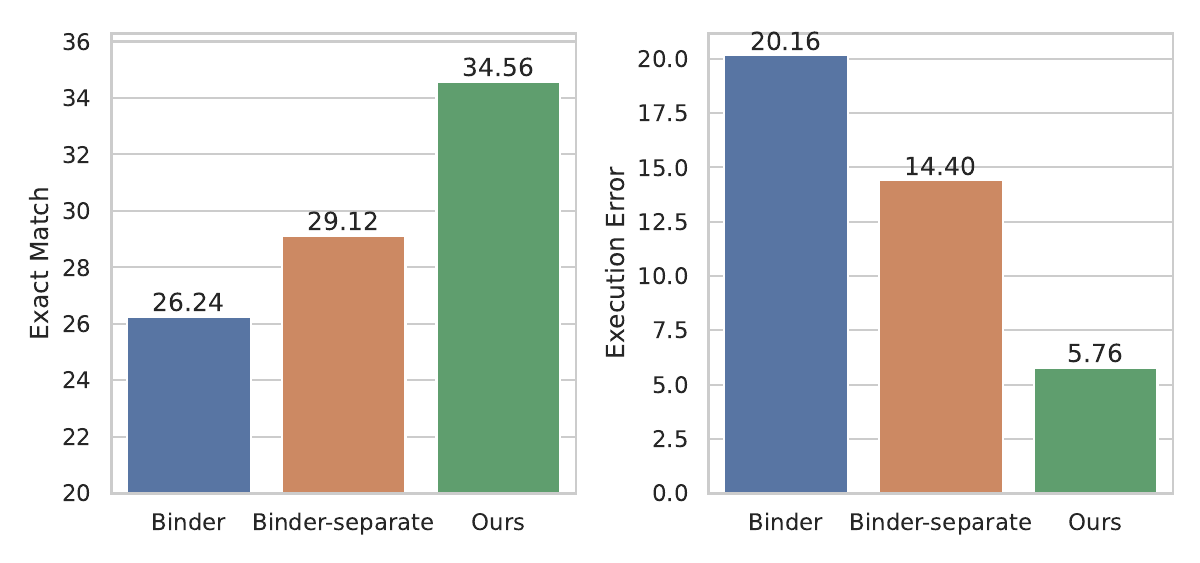}
    \vspace{-6mm}
    \caption{Comparison between our method and \texttt{Binder}.}
    \vspace{-4mm}
    \label{fig:binder-compare}
\end{figure}

\paragraph{Comparison with \texttt{Binder+CoT}.}
To further verify whether our augment-then-generate pipeline leads to easier and more accurate SQL generation over the best-performing baseline \texttt{Binder+CoT} (hereafter \texttt{Binder}), we compare the two methods on the subset of questions that are not solvable by pure SQL \citep{shi-etal-2020-potential}, which rely more on the integration of external knowledge. Figure \ref{fig:binder-compare} shows the EM and percentage of execution errors, where our method demonstrates a more pronounced improvement. To better pinpoint the cause of performance difference, we add a post-processing step for \texttt{Binder}, where we extract the LLM queries from the SQL statement generated by \texttt{Binder}, query LLMs for desired information and add it as a new column in the original table, and re-generate a standard SQL (without LLM queries) based on the augmented table. This variant (dubbed \texttt{Binder-separate}) improves the EM and reduces execution errors over \texttt{Binder}, which validates our hypothesis that combining LLM queries with SQL complicates the generation, leading to more syntax errors in generated programs. Notably, our method still incurs fewer execution errors than \texttt{Binder-separate}, which is likely due to the fact that our method generates more augmentations for the table, thus reducing the complexity of required SQL (see Appendix \ref{app:binder-example} for details and examples).

In Appendix \ref{app:chain-of-table}, we also compare our method with the concurrent work \texttt{Chain-of-Table} \citep{wang2024chainoftable}. Results show that our method achieves 1.85 higher EM when using GPT3.5 as the backbone LLM, demonstrating its effectiveness despite being simpler and not requiring sequential operations. Please refer to Appendix \ref{app:chain-of-table} for details.

\subsection{Closed-Domain Knowledge}
\label{subsec:exp-closed}
\paragraph{Datasets.}
For closed-domain setting, we experiment on \texttt{TATQA} \citep{zhu2021tat} and \texttt{FinQA} \citep{chen-etal-2021-finqa}. Questions in these two datasets are about a table and a financial report, and the answer often requires arithmetic operations in addition to table understanding ability. Since some questions are answerable with only the table or report, we filter the datasets to only include those that require both table and report (details in Table \ref{tab:datasets}).

\begin{figure}
    \centering
    \includegraphics[width=\linewidth]{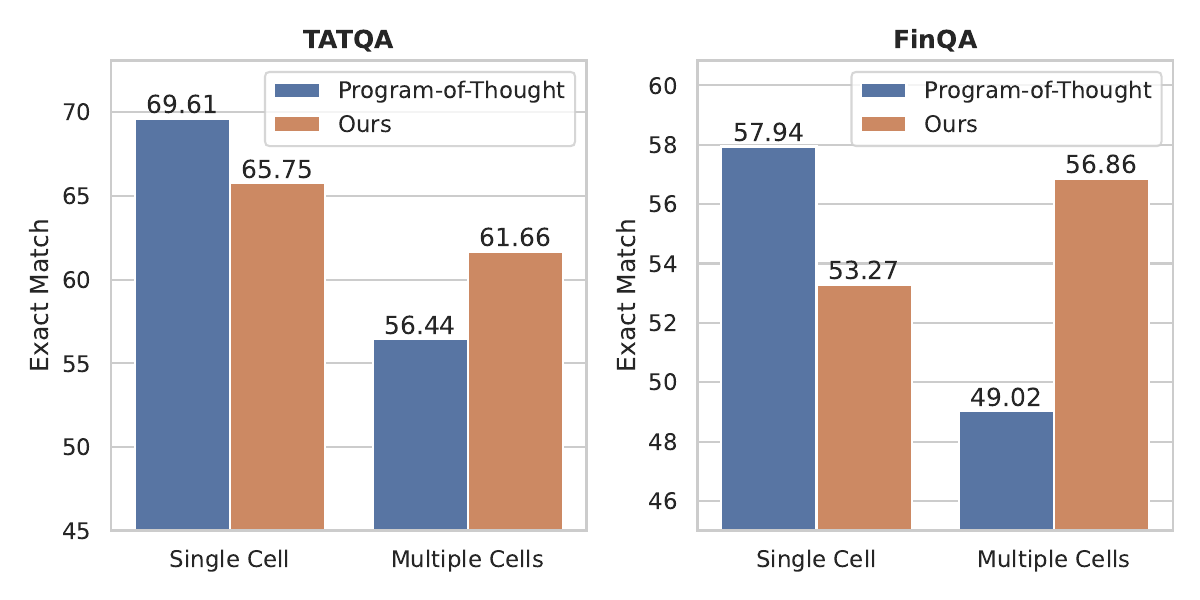}
    \vspace{-6mm}
    \caption{Performance decomposition by the number of table cells needed to answer the question.}
    \vspace{-3mm}
    \label{fig:complex_decompose}
\end{figure}

\paragraph{Results.}
Table \ref{tab:text} presents the results. \texttt{Binder} and \texttt{Dater} are not included because the original paper did not evaluate on these datasets and extension to this setting requires substantial modification. There are two observations. First, our method and \texttt{PoT} significantly outperform the other two baselines that do not involve program executions, which shows the benefits of leveraging programs when questions require arithmetic calculations. Second, although the input tables are much smaller, which is beneficial for \texttt{PoT}, our method is on par with \texttt{PoT} on \textsc{FinQA} and outperforms it by 2 EM on \textsc{TATQA}. A further performance breakdown by the number of table cells required to answer a question in Figure \ref{fig:complex_decompose} shows that our method is more effective on questions that require information from multiple cells, indicating that our method is more likely to generalize to complex questions. Furthermore, it is easier to locate and correct errors made by our method as it only requires inspection of the generated SQL queries, whereas \texttt{PoT} requires checking the whole table contents (see examples in Appendix \ref{app:pot-example}).

\input{tables/table-text}

%% file: tables/wikitq.tex
\begin{table}
\centering
\resizebox{0.75\linewidth}{!}
{
\begin{tabular}{lc}
\toprule
& \textbf{Test EM}
\\ 
\midrule
\texttt{End-to-End}
& 46.80
\\
\texttt{Table-CoT} \citep{chen-2023-large}
& 49.40
\\
\midrule
\texttt{PoT} \citep{chen2023program}
& 53.60
\\
\texttt{Dater} \citep{ye2023large}
& 48.40
\\
\texttt{Binder} \citep{cheng2023binding}
& 34.80
\\
\texttt{Binder+CoT}
& 54.10
\\
\texttt{Ours}
& \textbf{55.80}
\\
\bottomrule
\end{tabular}
}
\vspace{-2mm}
\caption{Exact match on \textsc{WikiTQ} test set.
Methods in the bottom panel involve program execution.}
\label{tab:wikitq}
\end{table}

%% file: tables/table-text.tex
\begin{table}
\centering
\resizebox{0.9\linewidth}{!}
{
\begin{tabular}{lcc}
\toprule
& \textbf{\textsc{TATQA}} & \textbf{\textsc{FinQA}}
\\ 
\midrule
\texttt{End-to-End}
& 35.50
& 34.18
\\
\texttt{Table-CoT} \citep{chen-2023-large}
& 34.91
& 39.87
\\
\texttt{PoT}\citep{chen2023program}
& 61.14
& \textbf{54.43}
\\
\texttt{Ours}
& \textbf{63.12}
& 53.80
\\
\bottomrule
\end{tabular}
}
\vspace{-2mm}
\caption{Exact match on \textsc{TATQA} and \textsc{FinQA}.}
\label{tab:text}
\end{table}

%% file: sections/05_conclusion.tex
In this paper, we propose a simple method that augments a given table by creating a new table that contains the information from external sources. The LLM then generates a SQL query over the two tables to answer the given question. Experiments show that our method outperforms or matches strong baselines on three table QA benchmarks.

%% file: sections/limitations.tex
There are several limitations in this work that need to be further improved. First, our framework relies on the LLM's ability to generate correct SQL statements. If the LLM has limited SQL generation ability, such as \texttt{Llama2} in Appendix \ref{app:chain-of-table}, the performance of our method will be affected. In addition, we only evaluate our method on integrating external knowledge from two different sources. The generalizability of our method to other knowledge sources remains to be assessed.

%% file: sections/06_appendix.tex
\section{Implementation Details}
\label{app:imp-details}

\input{tables/greedy_params}
\input{tables/sampling_params}
\input{tables/datasets}

For all methods on all three datasets, we use the greedy decoding for generation, \emph{i.e.,} temperature equals 0. Table \ref{tab:greedy-params} lists other generation parameters of our method. Table \ref{tab:datasets} shows the statistics of datasets used in this paper.

\subsection{Open-domain Knowledge}\label{app:open-domain}
For the open-domain knowledge setting on \textsc{WikiTQ}, since our method generates queries that will be asked for every single row in one or more columns, the constructed augmenting table will always have the same number of rows as the original table. For simplicity, we directly join the two tables based on the row index before feeding them to LLMs to generate the SQL statement in step 3. In other words, the newly constructed table is joined on the original table as additional columns, and the SQL statement will be generated based on the joined table. Figures \ref{fig:wikitq-augment-system-prompt} and \ref{fig:wikitq-augment-in-context} show the detailed instruction used for this step and a demonstration of the in-context example. For step 2, we use the same instruction and in-context examples as \citet{cheng2023binding} to query LLMs for required information. An example is shown in Figure \ref{fig:wikitq-query-in-context}. For step 3, we provide LLMs with in-context examples along with a one-sentence instruction, as illustrated in Figure \ref{fig:wikitq-sql-in-context}.

To select a subset of 1000 questions for evaluation, we select one question in every four questions of the first 4000 samples in \textsc{WikiTQ}, \emph{i.e.,} \texttt{dataset = dataset.select(range(0, 4000, 4))}. We use the evaluation code in \citet{cheng2023binding} to calculate the EM for all methods.

\subsection{Closed-domain Knowledge}\label{app:llm-hyperparam}
For the closed-domain setting on \textsc{TATQA} and \textsc{FinQA}, we feed both the text document and the table to the LLM to provide enough context. To save the inference cost, we merge steps 1 and 2 together such that the model analyzes the required additional information and then extracts them from the document in a single run. We instruct the model to extract information in a JSON format that can be easily organized into a table. Figures \ref{fig:tatqa-augment-system-prompt} and \ref{fig:tatqa-augment-in-context} show the detailed instruction used and a demonstration of the in-context example on \textsc{TATQA}, and Figures \ref{fig:finqa-augment-system-prompt} and \ref{fig:finqa-augment-in-context} show the same for \textsc{FinQA}. For step 3, we provide the original table and the newly constructed table if available to LLMs. Figures \ref{fig:tatqa-sql-in-context} and \ref{fig:finqa-sql-in-context} show a demonstration of the in-context examples on \textsc{TATQA} and \textsc{FinQA} respectively.

To select questions for evaluation, we only use those that require both the table and the document. Specifically, for \textsc{TATQA}, we select questions that have \texttt{answer\_from}$=$\texttt{table-text}, and for \textsc{FinQA}, we select those whose ground truth evidence contains at least one table row and one document sentence. We follow \citet{chen2023program} to calculate the EM.

\section{Comparison with \texttt{Chain-of-Table}}
\label{app:chain-of-table}

\input{tables/chain-of-table}

We additionally compare with \texttt{Chain-of-Table} \citep{wang2024chainoftable} on \textsc{WikiTQ}. Since their implementation is not available at the submission time of this paper, we use the same dataset and backbone LLMs as theirs and directly compare with the numbers reported in their paper. Specifically, we use \texttt{GPT-3.5-turbo-16k-0613} and \texttt{Llama2-13b-chat} \citep{touvron2023llama} as backbone LLMs and evaluate on the full test set of \textsc{WikiTQ}. Since their sequential operations require multiple queries for LLMs, we consider the majority vote of execution results from \e{N} SQL queries as our final prediction. To generate these SQL queries, we sample \e{m} different outputs for step 1 (\emph{i.e.,} \e{m} different augmentations), and for each augmentation, we sample \e{k} SQL queries. The total number of generated samples for each question is \e{m + \alpha m + mk}, where \e{\alpha} is the percentage of step 1 outputs that actually need additional information. Table \ref{tab:sample-params} lists the parameters for generation.

The results are shown in Table \ref{tab:chain-of-table}. As can be observed, our method outperforms \texttt{Chain-of-Table} and \texttt{Binder} when using GPT3.5 as the backbone LLM, despite using fewer LLM queries. When using Llama2, \texttt{Chain-of-Table} achieves better performance than \texttt{Binder} and our method. We hypothesize that the performance difference is due to the limited SQL generation ability of Llama2. An important difference is that \texttt{Chain-of-Table} feeds the final table to LLMs and directly asks LLMs to generate the answer, whereas \texttt{Binder} and our method prompt LLMs to generate SQL queries and execute to get the answer, which is affected more when the LLM has limited SQL generation ability. In fact, the generated SQL of our method contains $32.9\%$ of execution errors when using Llama2 as the LLM, compared to that of $8.7\%$ when using GPT3.5. However, our method still outperforms \texttt{Binder} on Llama2, demonstrating the benefits of our augment-then-generate pipeline.

\section{Additional Examples}
\subsection{Comparison with \texttt{Binder}}
\label{app:binder-example}
\begin{figure}[t]
    \centering
    \includegraphics[width=\linewidth]{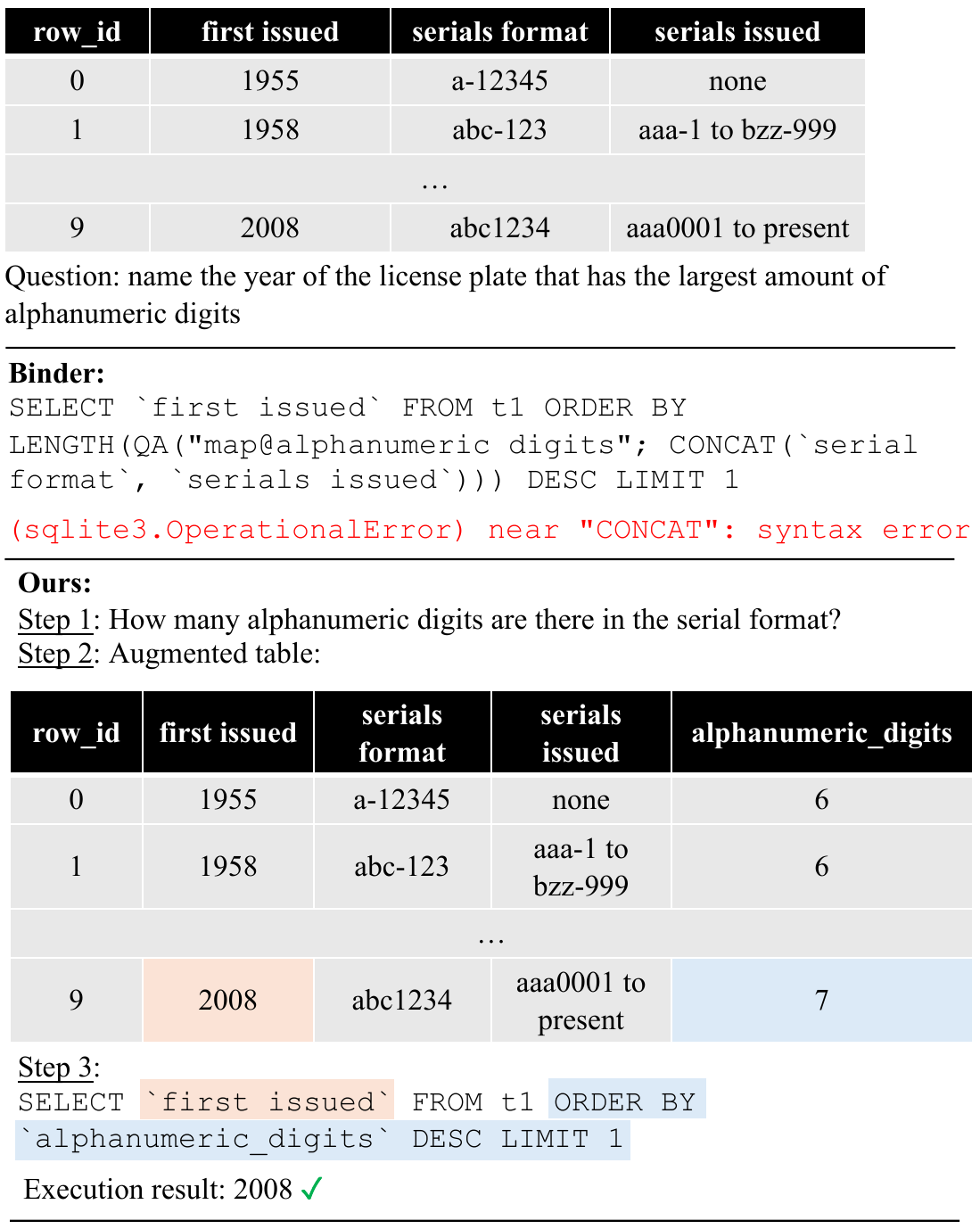}
    \caption{An example question in \textsc{WikiTQ}. \texttt{Binder} generates a SQL statement that queries LLMs for unsolvable parts. However, the statement leads to an execution error. Our method augments the table with an additional column and correctly generates a SQL statement to answer the question.}
    \vspace{-5mm}
    \label{fig:binder-failure-case1}
\end{figure}

\begin{figure}[t]
    \centering
    \includegraphics[width=\linewidth]{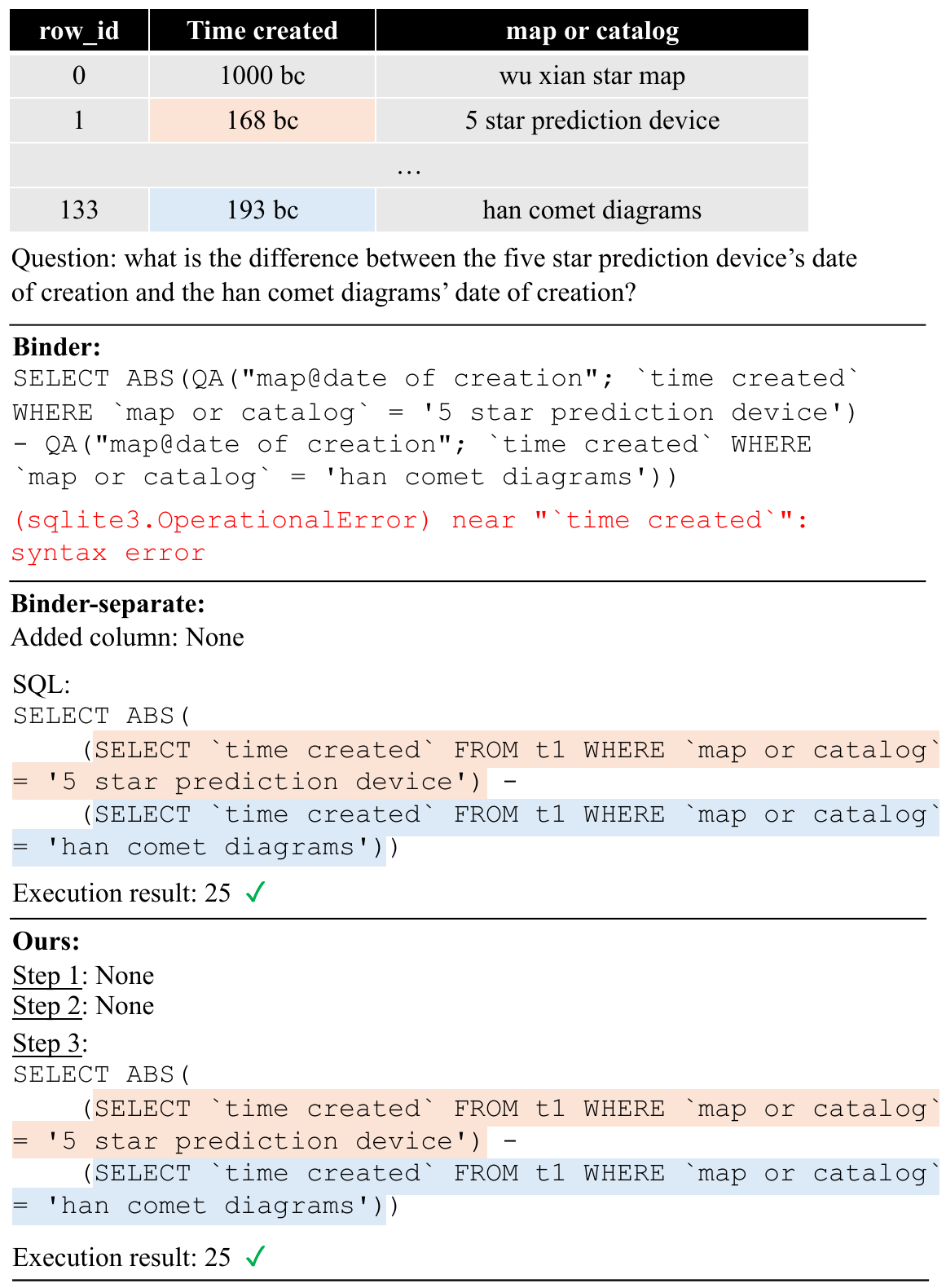}
    \caption{An example question in \textsc{WikiTQ}. \texttt{Binder} generates a SQL statement that queries LLMs. However, the statement leads to an execution error. Our method and \texttt{Binder-separate} correctly generate a pure SQL statement to answer the question.}
    \label{fig:binder-failure-case2}
\end{figure}

In this section, we elaborate on the comparison between our method, \texttt{Binder}, and \texttt{Binder-separate}. In Figure \ref{fig:binder-compare}, it can be observed that our method achieves better performance and exhibits fewer execution errors than \texttt{Binder}. Moreover, \texttt{Binder-separate}, which separates the SQL generation and LLM queries in \texttt{Binder}, reduces its execution errors, validating our hypothesis that integrating LLM queries in SQL generation could lead to more syntax errors. Figures \ref{fig:binder-failure-case1} and \ref{fig:binder-failure-case2} show two examples where \texttt{Binder} encounters execution errors when trying to generate a SQL statement with LLM queries, whereas our method and \texttt{Binder-separate} correctly generate SQL statements to answer the question.

\begin{figure}[t]
    \centering
    \includegraphics[width=\linewidth]{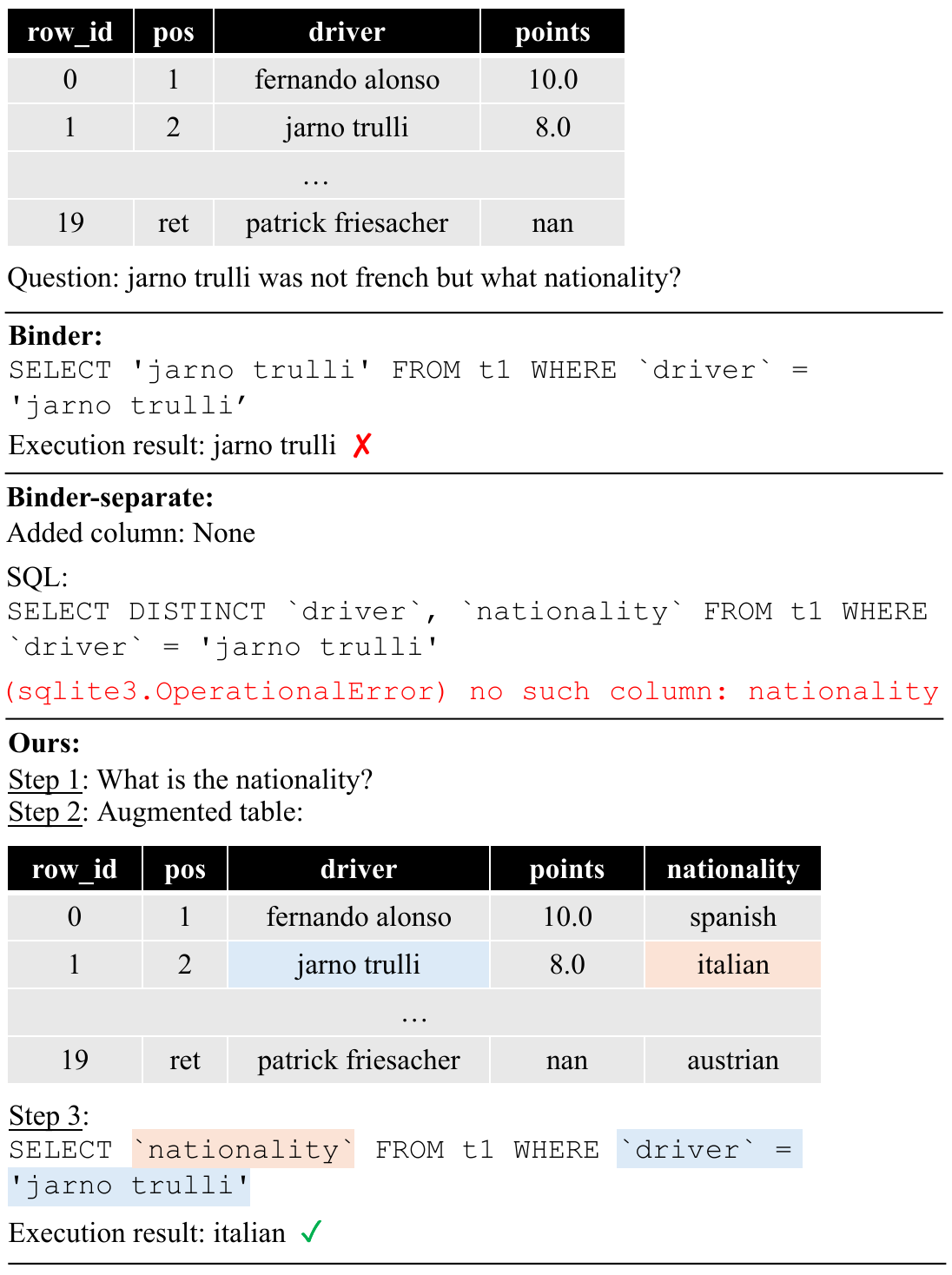}
    \caption{An example question in \textsc{WikiTQ}. \texttt{Binder} generates a SQL statement without syntax error, but it does not query LLMs for additional knowledge, resulting in a wrong answer. Since the original SQL statement generated by \texttt{Binder} does not augment the table with additional information, the question is not answerable using pure SQL, leading to an execution error in \texttt{Binder-separate}. Our method augments the table and correctly answers the question.}
    \label{fig:binder-separate-failure-case}
    \vspace{-5mm}
\end{figure}

Our method also incurs fewer execution errors than \texttt{Binder-separate}, which can be ascribed to the fact that our method generates more augmentations for the table, which significantly reduces the complexity of required SQL statements. Figure \ref{fig:binder-separate-failure-case} illustrates one such example, where \texttt{Binder-separate} gets errors because the required information is missing from the table, whereas our method correctly answers the question based on the augmented table. In fact, our method generates augmentations for $72.3\%$ of the questions, while \texttt{Binder} only includes LLM queries for $6.1\%$ of the questions, showing that our method also benefits the augmentation of additional information.

\subsection{Comparison with \texttt{PoT}}
\label{app:pot-example}
\begin{figure}[t]
    \centering
    \includegraphics[width=\linewidth]{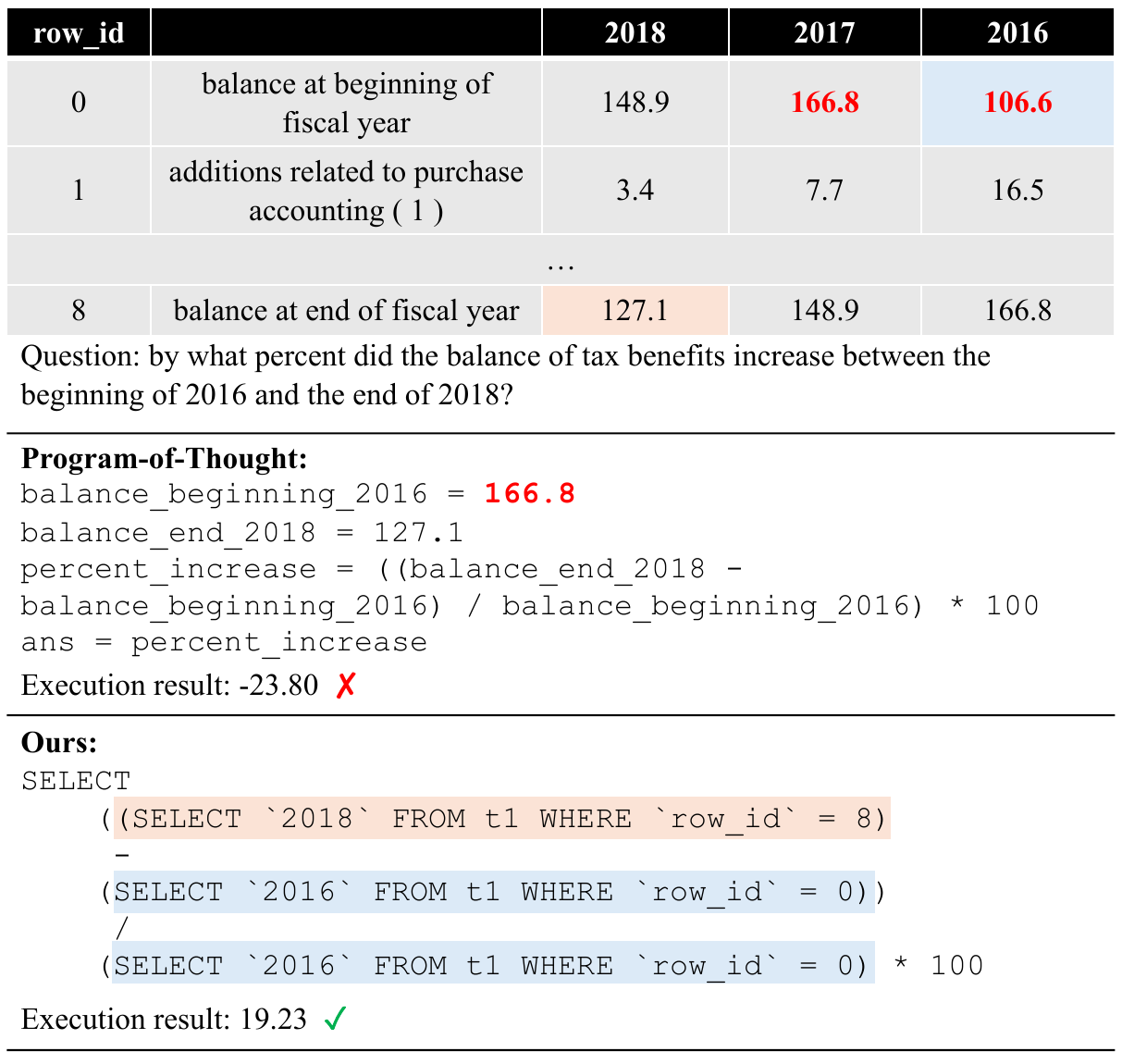}
    \caption{An example question in \textsc{FinQA} that requires two table cells to answer. \texttt{PoT} retrieves the wrong value (highlighted in \textcolor{red}{red}) from the table, despite generating a program with correct logic. Identifying the error requires looking into the table contents manually. Our method correctly selects the values and answers the question.}
    \label{fig:pot_example1}
    \vspace{-5mm}
\end{figure}

\begin{figure}[t]
    \centering
    \includegraphics[width=\linewidth]{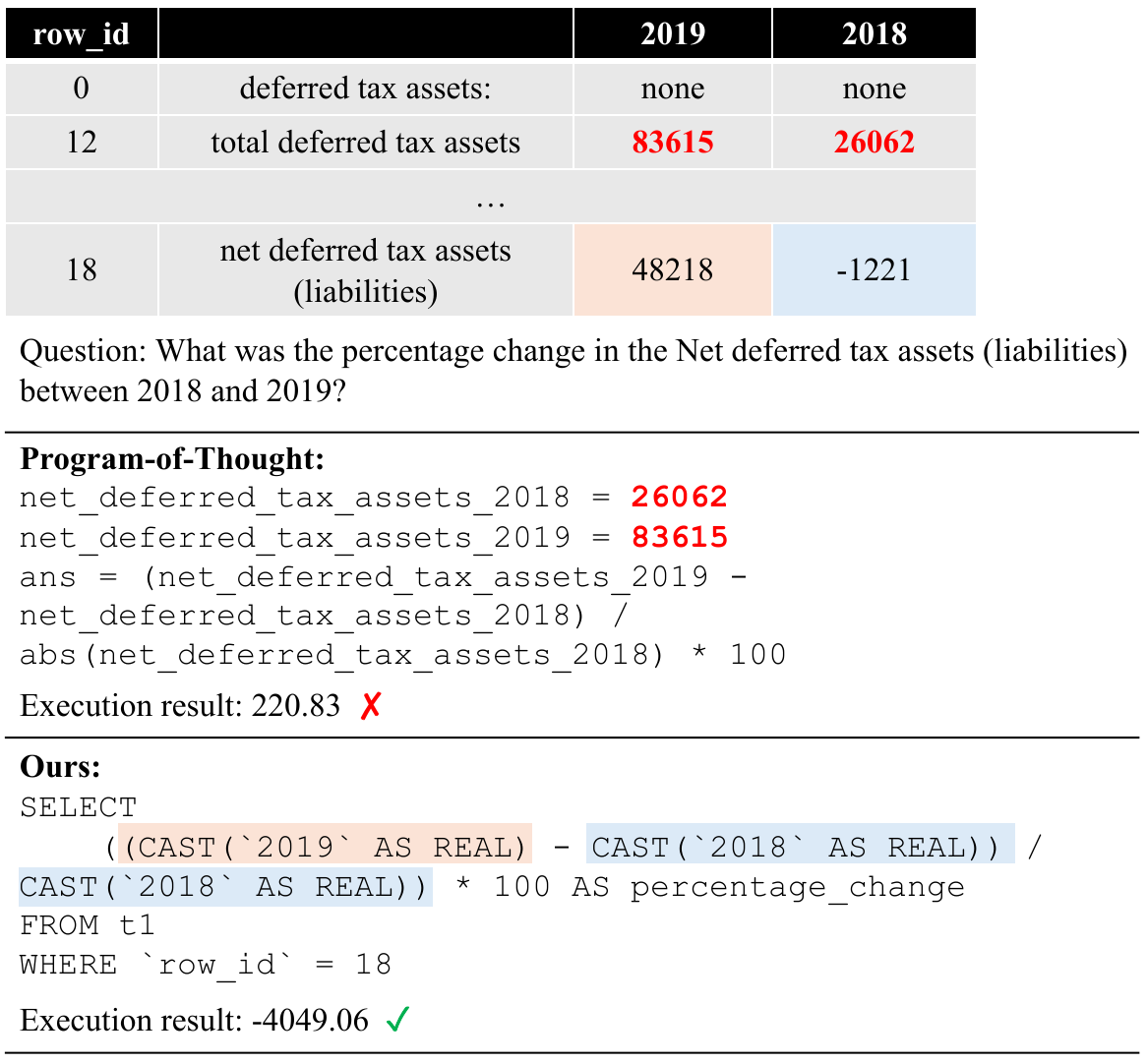}
    \caption{An example question in \textsc{TATQA} that requires two table cells to answer. \texttt{PoT} retrieves the wrong value (highlighted in \textcolor{red}{red}) from the table, despite generating a program with correct logic. Our method correctly selects the values and answers the question.}
    \label{fig:pot_example2}
    \vspace{-5mm}
\end{figure}

We now provide more examples for the comparison between our method and \texttt{PoT}. Based on Figure \ref{fig:complex_decompose}, our method is more effective on questions that require multiple table cells for the answer. Figures \ref{fig:pot_example1} and \ref{fig:pot_example2} show two such examples, where our method selects the correct values from the table to perform calculations, but \texttt{PoT} retrieves wrong values from the table, despite generating programs with correct logic. According to \citet{chen2023program}, this type of value grounding errors take up 47$\%$ of the errors made by \texttt{PoT}. Moreover, correcting these errors requires manual efforts to look into the contents of the table, which is time-consuming when the table is large.

\begin{figure}[t]
    \centering
    \includegraphics[width=\linewidth]{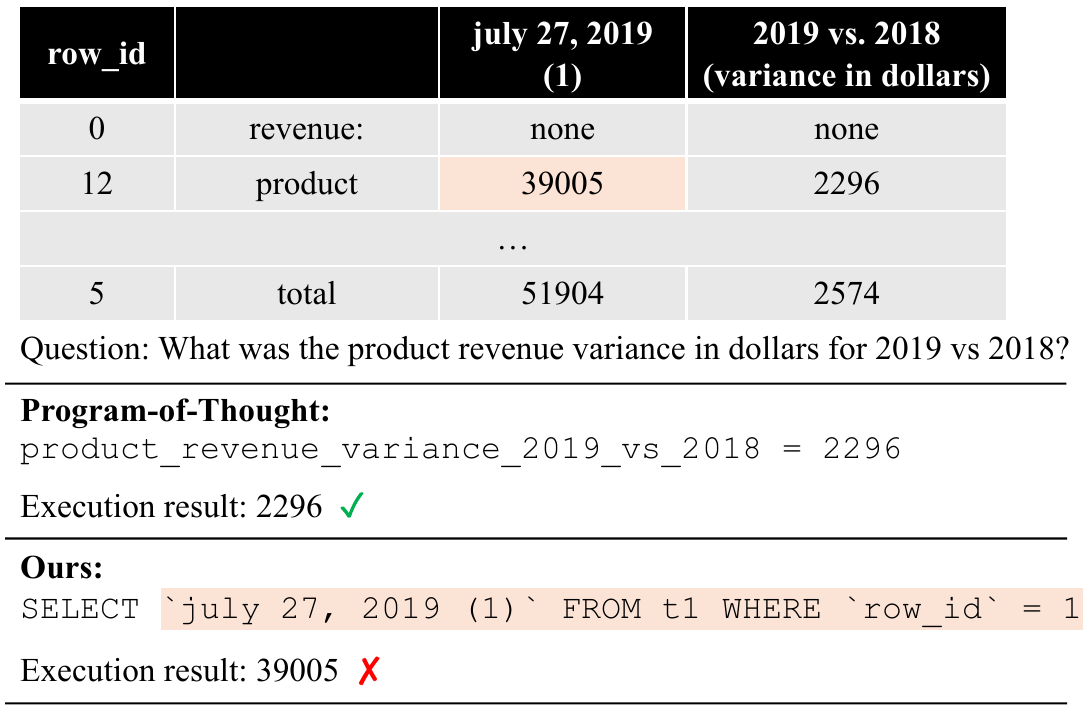}
    \caption{An example question in \textsc{TATQA} that requires a single table cell to answer. \texttt{PoT} correctly retrieves the value from the table. Our method mistakenly selects the value. However, the error is easy to be spotted and corrected by inspecting the SQL statement.}
    \label{fig:pot_example3}
    \vspace{-5mm}
\end{figure}

On the contrary, Figure \ref{fig:pot_example3} shows an example question that only requires a single cell from the table. \texttt{PoT} correctly selects the answer but our method selects the value in the wrong column. However, correcting this error requires only manual inspection of the generated SQL statement, which is much more efficient than checking the whole table contents.

\begin{figure*}[t]
\begin{tcolorbox}
\begin{lstlisting}[style=text]
Task Description:
Your task is to prepare a table for SQL query generation in order to answer a specific question. This may require modifying the table by adding extra columns. These new columns are created based on natural language questions, with each question applied individually to every row in the existing columns. The goal is to transform existing data into a format that's suitable for SQL operations, or to incorporate additional information into the table.

Procedure:
1. Evaluate the Table and Question: Assess if the table, in its current format, is suitable for generating a SQL query to answer the given question.
2. Determine Additional Columns:
   - If the table is already suitable for the SQL query, simply output "None"
   - If the table requires modifications, identify and define the necessary changes. Specifically, add new columns where each row's value is derived from a natural language question applied to the relevant columns. Use the format:
   `new_column` = @("question"; [relevant_columns]),
   where `question` is the question asked for each row to derive the new column's contents, and `relevant_columns` are the existing columns that provide the information needed for the question.

Response Format:
Begin your response with "Transformation:" and include:
- Solution outline: Describe a step-by-step reasoning chain of how to answer the question.
- Further analysis: Determine if modifications are required for each step.
- Final output: List each required additional column in the specified format, each on a new line. If no modifications are needed, output "None".

\end{lstlisting}
\end{tcolorbox}
\caption{System prompt used for augmentation generation (Step 1) on \textsc{WikiTQ}.}
\label{fig:wikitq-augment-system-prompt}
\end{figure*}

\begin{figure*}[t]
\begin{tcolorbox}
\begin{lstlisting}[style=demo]
Title: 2007 New Orleans Saints season
CREATE TABLE t1(
    row_id int,
    date text,
    game site text,
    result/score text)
/*
3 example rows:
SELECT * FROM t1 LIMIT 3;
row_id    date         game site                result/score
0         2007-9-6     rca dome                 l 41-10
1         2007-9-16    raymond james stadium    l 31-14
2         2007-9-24    louisiana superdome      l 31-14
*/

Q: what number of games were lost at home?
Transformation: 
Solution outline:
1. Find the losing games.
2. Find the games at home.
3. Count the number of games that satisfy both conditions.
Further analysis:
For step 1, we need information in `result/score` column. We need to parse if it's a win or loss. We will add a column called `is_loss`.
For step 2, we need information in `game site` column. We need additional information on whether it's a home game or not. We will add a column called `is_home_game`.
Step 3 can be done with a SQL query.
Final output:
`is_loss` = @("Is it a loss?"; [result/score])
`is_home_game` = @("Is it the home court of New Orleans Saints?"; [game site])
\end{lstlisting}
\end{tcolorbox}
\caption{A demonstration of the in-context example used for augmentation generation (Step 1) on \textsc{WikiTQ}.}
\label{fig:wikitq-augment-in-context}
\end{figure*}

\begin{figure*}[t]
\begin{tcolorbox}
\begin{lstlisting}[style=demo]
Give a database as shown below:
Table: 1963 International Gold Cup
/*
row_id  driver
0    jim clark
1    richie ginther
2    graham hill
3    jack brabham
4    tony maggs
*/
Q: Answer question "What is his/her country?" row by row.
Output:
/*
row_id  driver
0    jim clark        scotland
1    richie ginther   united states
2    graham hill      england
3    jack brabham     australia
4    tony maggs       south africa
*/
\end{lstlisting}
\end{tcolorbox}
\caption{A demonstration of the in-context example used for querying additional information (Step 2) from LLMs on \textsc{WikiTQ}.}
\label{fig:wikitq-query-in-context}
\end{figure*}

\begin{figure*}
\begin{tcolorbox}
\begin{lstlisting}[style=demo]
Read the following table and write a SQL query to answer the question:
Title: 2007 New Orleans Saints season
CREATE TABLE t1(
    row_id int,
    date text,
    game site text,
    result/score text,
    is_loss text,
    is_home_game text)
/*
3 example rows:
SELECT * FROM t1 LIMIT 3;
row_id    date         game site                result/score    is_loss    is_home_game
0         2007-9-6     rca dome                 l 41-10         yes        no
1         2007-9-16    raymond james stadium    l 31-14         yes        no
2         2007-9-24    louisiana superdome      l 31-14         yes        yes
*/

Q: what number of games were lost at home?
SQL: To answer the question, we need following steps:
1. Find the losing games by `is_loss` column.
2. Find the games at home by `is_home_game` column.
3. Count the number of games that satisfy both conditions.
Final SQL query:
```
SELECT COUNT(*) FROM t1 WHERE `is_loss` = 'yes' AND `is_home_game` = 'yes'
```
\end{lstlisting}
\end{tcolorbox}
\caption{A demonstration of the in-context example used for SQL generation (Step 3) on \textsc{WikiTQ}.}
\label{fig:wikitq-sql-in-context}
\end{figure*}

\begin{figure*}[t]
\begin{tcolorbox}
\begin{lstlisting}[style=text]
Task Description:
You are tasked with analyzing a provided table and an accompanying report to answer a specific question. This involves assessing whether the table contains all necessary information for answering the question. If additional information is needed, you must extract this from the report and create a supplementary table. Your primary focus is on the analysis and information extraction process, which will facilitate in forming a SQL query to answer the question.

Procedure:
1. Assess the Given Table and Question: Determine whether the provided table contains all the required information to answer the question.
2. Extract Information for Additional Table Creation:
   - If the existing table is sufficient, simply output "None"
   - If the existing table lacks essential information, extract the required data from the report in the following JSON format: `{"column_name": [value1, ...], ...}`

Each example is given in the following structure:
- Report: Contents of the report that may contain additional information.
- Tables: Contents of the table, with columns separated by " | " and rows by "\n".
- Question: The specific question that needs to be answered.

Response Format:
Begin your response with "Analysis:" and include:
- Solution outline: Describe the step-by-step outline for answering the question.
- Further analysis: Determine whether each step's information is available in the existing table or needs to be extracted from the report.
- Final output: Extract necessary information from the report in JSON format as described above; if no additional information is needed, output "None".

Notes:
- You may extract information with any number of columns and rows. However, all columns should have the same number of values.
- Make the JSON self-explanatory. Use descriptive column names, add context where needed, and include units in column names to prevent ambiguity.
- Avoid creating columns with empty or NaN values.
\end{lstlisting}
\end{tcolorbox}
\caption{System prompt used for constructing augmenting table (Steps 1 and 2) on \textsc{TATQA}.}
\label{fig:tatqa-augment-system-prompt}
\end{figure*}

\begin{figure*}[t]
\begin{tcolorbox}
\begin{lstlisting}[style=demo]
Report:
NOTE 5 - PROPERTY AND EQUIPMENT
The Company owned equipment recorded at cost, which consisted of the following as of December 31, 2019 and 2018:
Depreciation expense was $80,206 and $58,423 for the years ended December 31, 2019 and 2018, respectively
Tables:
row_id | filledcolumnname | 2019 | 2018
0 | computer equipment | 137763 | 94384
1 | furniture and fixtures | 187167 | 159648
2 | subtotal | 324930 | 254032
3 | less accumulated depreciation | 148916 | 104702
4 | property and equipment, net | 176014 | 149330

Question: What is the ratio of depreciation expense to accumulated depreciation of property and equipment in 2019?
Analysis:
Solution outline:
1. Find the amount of depreciation expense and accumulated depreciation of property and equipment in 2019.
2. Calculate the ratio.
Further analysis:
For step 1, the accumulated depreciation is mentioned in the table in row 3. But the depreciation expense is missing from the table. So we need to extract it from the report.
Step 2 can be done with a SQL query.
Final output:
{"depreciation_expense_2019": ["$80,206"]}
\end{lstlisting}
\end{tcolorbox}
\caption{A demonstration of the in-context example used for constructing augmenting table (Steps 1 and 2) on \textsc{TATQA}.}
\label{fig:tatqa-augment-in-context}
\end{figure*}

\begin{figure*}[t]
\begin{tcolorbox}
\begin{lstlisting}[style=demo]
Report:
NOTE 5 - PROPERTY AND EQUIPMENT The Company owned equipment recorded at cost, which consisted of the following as of December 31, 2019 and 2018: Depreciation expense was $80,206 and $58,423 for the years ended December 31, 2019 and 2018, respectively
Tables:
CREATE TABLE t1(
    row_id int,
    filledcolumnname text,
    2019 int,
    2018 int)
/*
All rows of the table:
SELECT * FROM t1;
row_id	filledcolumnname	2019	2018
0	computer equipment	137763	94384
1	furniture and fixtures	187167	159648
2	subtotal	324930	254032
3	less accumulated depreciation	148916	104702
4	property and equipment, net	176014	149330
*/

CREATE TABLE t2(
    row_id int,
    depreciation_expense_2019 int)
/*
All rows of the table:
SELECT * FROM t2;
row_id	depreciation_expense_2019
0	80206
*/

Q: What is the ratio of depreciation expense to accumulated depreciation of property and equipment in 2019?
SQL: Reasoning process:
We need following steps to answer the question:
1. Get the depreciation expense in 2019 from t2.
2. Get the accumulated depreciation in 2019 from t1, which is in row 3.
3. Calculate the ratio.
Final SQL query:
```
SELECT 
    (SELECT `depreciation_expense_2019` FROM t2 WHERE `row_id` = 0) / 
    CAST((SELECT `2019` FROM t1 WHERE `row_id` = 3) AS REAL) 
    AS depreciation_ratio
FROM t1
LIMIT 1
```
Units: ""
\end{lstlisting}
\end{tcolorbox}
\caption{A demonstration of the in-context example used for SQL generation (Step 3) on \textsc{TATQA}.}
\label{fig:tatqa-sql-in-context}
\end{figure*}

\begin{figure*}[t]
\begin{tcolorbox}
\begin{lstlisting}[style=text]
Task Procedure:
1. Assess the Given Table and Question: Determine whether the provided table contains all the required information to answer the question.
2. Extract Missing Information from Report:
   - If the existing table is sufficient, simply output "None"
   - If the existing table lacks essential information, extract the required data from the report in the following JSON format: `{"column_name": [value1, ...], ...}`

Each example is given in the following structure:
- Report: Contents of the report that may contain additional information.
- Tables: Contents of the table, with columns separated by " | " and rows by "\n".
- Question: The specific question that needs to be answered.

Response Format:
Begin your response with "Analysis:" and include:
- Solution formula: Write a formula to calculate the answer.
- Further analysis: Determine for each variable in the formula whether it is available in the table or needs to be extracted from the report.
- Final output: For variables not in the table, extract them from report in JSON format as described above; if all variables are in the table, output "None".

Notes:
- Make the JSON self-explanatory. Use descriptive column names and include units in column names to prevent ambiguity.
\end{lstlisting}
\end{tcolorbox}
\caption{System prompt used for constructing augmenting table (Steps 1 and 2) on \textsc{FinQA}.}
\label{fig:finqa-augment-system-prompt}
\end{figure*}

\begin{figure*}[t]
\begin{tcolorbox}
\begin{lstlisting}[style=demo]
Report:
purchases of equity securities 2013 during 2014 , we repurchased 33035204 shares of our common stock at an average price of $ 100.24 .
[b] effective january 1 , 2014 , our board of directors authorized the repurchase of up to 120 million shares of our common stock by december 31 , 2017 .
Tables:
row_id | period | total number ofsharespurchased[a] | averageprice paidpershare | total number of sharespurchased as part of apublicly announcedplan or program [b] | maximum number ofshares that may yetbe purchased under the planor program [b]
0 | oct . 1 through oct . 31 | 3087549 | 107.59 | 3075000 | 92618000
1 | nov . 1 through nov . 30 | 1877330 | 119.84 | 1875000 | 90743000
2 | dec . 1 through dec . 31 | 2787108 | 116.54 | 2786400 | 87956600
3 | total | 7751987 | 113.77 | 7736400 | n/a

Question: what percent of the share repurchases were in the fourth quarter?
Analysis:
Solution formula:
share_repurchase_fourth_quarter / share_repurchase_whole_year
Further analysis:
share_repurchase_fourth_quarter is in row 3 of the table
share_repurchase_whole_year is not in the table, so we need to extract it from the report
Final output:
{"share_repurchase_whole_year": [33035204]}
\end{lstlisting}
\end{tcolorbox}
\caption{A demonstration of the in-context example used for constructing augmenting table (Steps 1 and 2) on \textsc{FinQA}.}
\label{fig:finqa-augment-in-context}
\end{figure*}

\begin{figure*}[t]
\begin{tcolorbox}
\begin{lstlisting}[style=demo]
Report:
purchases of equity securities 2013 during 2014 , we repurchased 33035204 shares of our common stock at an average price of $ 100.24 .
[b] effective january 1 , 2014 , our board of directors authorized the repurchase of up to 120 million shares of our common stock by december 31 , 2017 .
Tables:
CREATE TABLE t1(
    row_id int,
    period text,
    total number ofsharespurchased[a] int,
    averageprice paidpershare real,
    total number of sharespurchased as part of apublicly announcedplan or program [b] int,
    maximum number ofshares that may yetbe purchased under the planor program [b] text)
/*
All rows of the table:
SELECT * FROM t1;
row_id	period	total number ofsharespurchased[a]	averageprice paidpershare	total number of sharespurchased as part of apublicly announcedplan or program [b]	maximum number ofshares that may yetbe purchased under the planor program [b]
0	oct . 1 through oct . 31	3087549	107.59	3075000	92618000
1	nov . 1 through nov . 30	1877330	119.84	1875000	90743000
2	dec . 1 through dec . 31	2787108	116.54	2786400	87956600
3	total	7751987	113.77	7736400	n/a
*/

CREATE TABLE t2(
    row_id int,
    share_repurchase_whole_year int)
/*
All rows of the table:
SELECT * FROM t2;
row_id	share_repurchase_whole_year
0	33035204
*/

Q: what percent of the share repurchases were in the fourth quarter?
SQL: 
Solution formula:
share_repurchase_fourth_quarter / share_repurchase_whole_year
Further analysis:
share_repurchase_fourth_quarter is in row 3, column `total number ofsharespurchased[a]` of t1
share_repurchase_whole_year is in row 0, column `share_repurchase_whole_year` of t2
Final SQL query:
```
SELECT 
    CAST((SELECT `total number ofsharespurchased[a]` FROM t1 WHERE `row_id` = 3) AS REAL) / 
    (SELECT `share_repurchase_whole_year` FROM t2 WHERE `row_id` = 0) * 100
```
\end{lstlisting}
\end{tcolorbox}
\caption{A demonstration of the in-context example used for SQL generation (Step 3) on \textsc{FinQA}.}
\label{fig:finqa-sql-in-context}
\end{figure*}

%% file: tables/greedy_params.tex
\begin{table}
\centering
\resizebox{0.85\linewidth}{!}
{
\begin{tabular}{lccc}
\toprule
& \textbf{\textsc{WikiTQ}} & \textbf{\textsc{TATQA}} & \textbf{\textsc{FinQA}}
\\ 
\midrule
top\_p & 1.0 & 1.0 & 1.0
\\
max\_output\_tokens & 512 & 512 & 512
\\
num\_shots & 8 & 8 & 4
\\
\bottomrule
\end{tabular}
}
\caption{Parameters for our greedy generation (sections \ref{subsec:exp-open} and \ref{subsec:exp-closed}).}
\label{tab:greedy-params}
\end{table}

%% file: tables/sampling_params.tex
\begin{table}
\centering
\resizebox{\linewidth}{!}
{
\begin{tabular}{lcccc}
\toprule
& \multicolumn{2}{c}{\textbf{\texttt{GPT3.5}}} & \multicolumn{2}{c}{\textbf{\texttt{Llama2}}}
\\ 
\cmidrule(lr){2-3}
\cmidrule(lr){4-5}
& Augmentation & SQL & Augmentation & SQL \\
& generation & generation & generation & generation \\
\midrule
temperature & 0.6 & 0.4 & 0.8 & 0.4
\\
top\_p & 1.0 & 1.0 & 1.0 & 1.0
\\
sampling\_n & 3 & 2 or 4 & 4 & 3 or 4
\\
max\_output\_tokens & 512 & 512 & 256 & 256
\\
num\_shots & 8 & 8 & 8 & 8 \\
\bottomrule
\end{tabular}
}
\caption{Generation parameters for our ensemble model on \textsc{WikiTQ} (Appendix \ref{app:chain-of-table}). Augmentation generation and SQL generation correspond to the step 1 and 3 in our method.}
\label{tab:sample-params}
\end{table}

%% file: tables/datasets.tex
\begin{table*}[!t]
\centering
\resizebox{0.8\linewidth}{!}
{
\begin{tabular}{lcccc}
\toprule
& \textbf{\textsc{WikiTQ}} & \textbf{\textsc{WikiTQ} SQL unsolvable} & \textbf{\textsc{TATQA}} & \textbf{\textsc{FinQA}}
\\ 
\midrule
\# questions
& 1000
& 625
& 507
& 158
\\
Split
& Test
& Dev
& Dev
& Test
\\
\# table rows
& 26.1
& 28.0
& 9.7
& 6.8
\\
\# table tokens
& 571.7
& 685.7
& 119.1
& 86.2
\\
Knowledge source \e{S}
& LLMs
& LLMs
& Document
& Document
\\
\bottomrule
\end{tabular}
}
\caption{Summary of the datasets used in this paper.}
\label{tab:datasets}
\end{table*}

%% file: tables/chain-of-table.tex
\begin{table}
\centering
\resizebox{0.8\linewidth}{!}
{
\begin{tabular}{lcc}
\toprule
& \textbf{\# generated samples} & \textbf{EM}
\\
\midrule
\multicolumn{3}{c}{\textbf{\texttt{GPT3.5}}} \\
\midrule
\texttt{Binder} & 50 & 56.74
\\
\texttt{Chain-of-Table} & $\le25$ & 59.94
\\
\texttt{Ours} (6 SQLs) & 11.4 & 61.05
\\
\texttt{Ours} (12 SQLs) & 17.4 & \textbf{61.79}
\\
\midrule
\multicolumn{3}{c}{\textbf{\texttt{Llama2}}} \\
\midrule
\texttt{Binder} & 50 & 30.92
\\
\texttt{Chain-of-Table} & $\le25$ & \textbf{42.61}
\\
\texttt{Ours} (12 SQLs) & 19.82 & 34.00
\\
\texttt{Ours} (16 SQLs) & 23.82 & 35.34
\\
\bottomrule
\end{tabular}
}
\caption{Exact match on full \textsc{WikiTQ} test set. $\#$ generated samples denotes the total number of generated samples to answer one question.}
\label{tab:chain-of-table}
\end{table}